\documentclass[final,3p,times,twocolumn]{elsarticle}

\usepackage{amssymb}
\usepackage{url}
\usepackage{amsmath}
\usepackage{booktabs}       
\usepackage{amsfonts}       
\usepackage{nicefrac}       
\usepackage{microtype}      
\usepackage{xcolor}         
\usepackage{multirow}
\usepackage{graphicx}
\usepackage{wrapfig}
\usepackage{bm}
\usepackage[utf8]{inputenc}
\usepackage{hyperref}

\AtEndPreamble{
    \usepackage[capitalize]{cleveref}
    \crefname{section}{Sec.}{Secs.}
    \Crefname{section}{Section}{Sections}
    \Crefname{table}{Table}{Tables}
    \crefname{table}{Tab.}{Tabs.}
}

\journal{Computers \& Security}

\begin{document}

\begin{frontmatter}

\title{Harmonizing Feature Maps: A Graph Convolutional Approach for Enhancing Adversarial Robustness}


\author[a]{Kejia Zhang}
\author[a]{Juanjuan Weng}
\author[a]{Junwei Wu}
\author[a]{Guoqing Yang}
\author[a,b]{Shaozi Li}
\author[a]{Zhiming Luo}

\affiliation[a]{organization={Department of Artificial Intelligence, Xiamen University},
            city={Xiamen},
            postcode={361005}, 
            state={Fujian},
            country={China}
}

\affiliation[b]{organization={Fujian Key Laboratory of Big Data Application and Intellectualization for Tea Industry, Wuyi University},
city={Wuyishan},
            postcode={354300}, 
            state={Fujian},
            country={China}}

\begin{abstract}
The vulnerability of Deep Neural Networks to adversarial perturbations presents significant security concerns, as the imperceptible perturbations can contaminate the feature space and lead to incorrect predictions.
Recent studies have attempted to calibrate contaminated features by either suppressing or over-activating particular channels. 
Despite these efforts, we claim that adversarial attacks exhibit varying
disruption levels across individual channels.
Furthermore, we argue that harmonizing feature maps via graph and employing graph convolution can calibrate contaminated features.
To this end, we introduce an innovative plug-and-play module called Feature Map-based Reconstructed Graph Convolution (FMR-GC). 
FMR-GC harmonizes feature maps in the channel dimension to reconstruct the graph, then employs graph convolution to capture neighborhood information, effectively calibrating contaminated features. 
Extensive experiments have demonstrated the superior performance and scalability of FMR-GC. Moreover, our model can be combined with advanced adversarial training methods to considerably enhance robustness without compromising the model's clean accuracy.
\end{abstract}



\begin{keyword}
Adversarial Defense \sep Neural Network Robust \sep Graph Convolution Networks


\end{keyword}

\end{frontmatter}


\section{Introduction}
Deep Neural Networks (DNNs) have achieved significant progress in various computer vision tasks, \textit{e.g.}, image classification~\citep{wang2017residual}, image segmentation~\citep{minaee2021image}, and object detection~\citep{Object_detection_tnnls}. However, recent studies~\citep{long2022survey, zhao2024evaluating,szegedy2013intriguing} revealed that DNNs are vulnerable to adversarial attacks, where adding imperceptible perturbations to natural inputs can lead DNNs to make incorrect predictions. Therefore, the development of reliable and robust DNNs is essential to defend against such potential security threats.

\begin{figure}[t]
    \begin{center}
    \includegraphics[width=\linewidth]{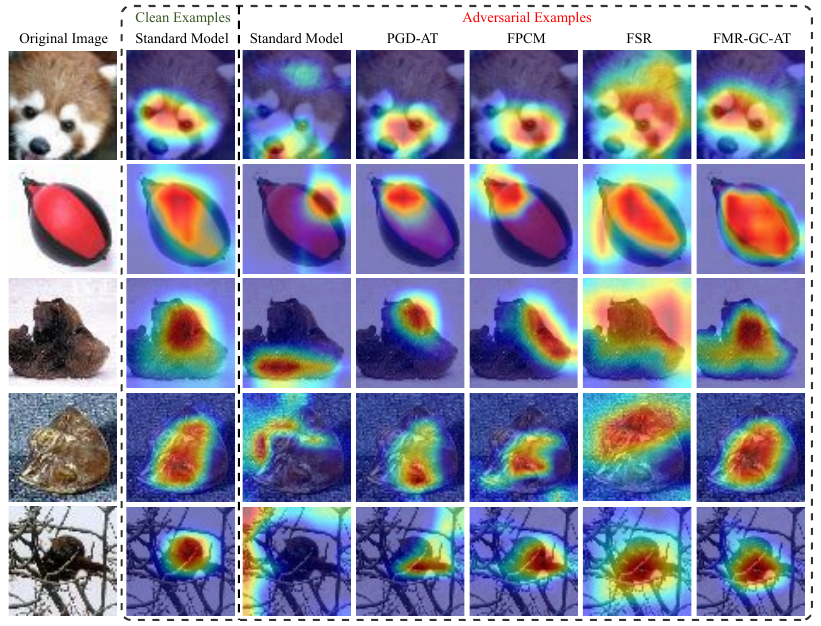}
    \end{center}
    \caption{Visualization of feature activation maps via various training methods utilizing Grad-CAM~\citep{selvaraju2017grad}. The 'standard model' denotes a model trained on clean examples. 
    Left-to-right: original image, the feature activation map of clean examples on the standard model, and the feature activation maps of adversarial examples on the standard model, PGD-AT, FPCM, FSR, and FMR-GC-AT.}
    \label{CAM_with_t_column}
\end{figure}

Adversarial training (AT) has been commonly used as an effective strategy for improving model robustness~\citep{PGD_AT, TRADES, AWP}, which employs adversarial examples (AEs) as training data during the training process. 
\citet{Xie_2019_CVPR} illustrated that subtle adversarial perturbations at the pixel level can significantly disrupt the feature space, shifting the attention area of the normal model (as depicted in Column 3 of \Cref{CAM_with_t_column}). 
However, conventional AT methods still struggle to effectively realign the model's attention back to a normal state when under adversarial attack (as depicted in Column 4 of \Cref{CAM_with_t_column}).
To address this challenge, recent feature-based AT methods conduct a feature calibration to reduce the impact of the disrupted features and enable correct predictions. For example, the Frequency Preference Control Module (FPCM) introduced by \citep{FPCM} is designed to calibrate high-frequency signals by employing a low-pass filter for suppression.
Besides, \citet{kim2023feature} propose a Feature Separation and Recalibration (FSR) method, which recalibrates non-robust activations to explore potential discriminative features. 


To further evaluate the effectiveness of FPCM~\citep{FPCM} and FSR~\citep{kim2023feature}, we analyzed their feature attention maps during network inference, as illustrated in \Cref{CAM_with_t_column}. It is evident that FPCM deactivates useful feature activation regions compared to the baseline state (refer to Column 5 in \Cref{CAM_with_t_column}). This can be attributed to the low-pass filtering in FPCM, which suppresses the activation of certain channels. In contrast, FSR encounters the issue of feature activation spreading to semantically irrelevant areas (refer to Column 6 in \Cref{CAM_with_t_column}). This is primarily because the reuse of non-robust features results in excessive activation.

On the other aspect, recent studies~\citep{Zhang_2022_CVPR, Wu_2020_CVPR} have revealed that adversarial attacks manifest varying levels of feature activation across different channels during network inference.  
These findings suggest that some channels may be significantly affected by adversarial attacks, while others may encounter minimal influences. 
Moreover, \citet{yu2020graph, li2021cross} have demonstrated that graph convolution, which leverages contextual information from neighboring nodes, can mitigate node contamination. 
Building upon the aforementioned discovery, we question whether it is feasible to use graph convolution to calibrate the contaminated features of adversarial samples by exploring inter-feature relationships at the channel level. 


\begin{figure}[t]
    \begin{center}
    \includegraphics[width=\linewidth]{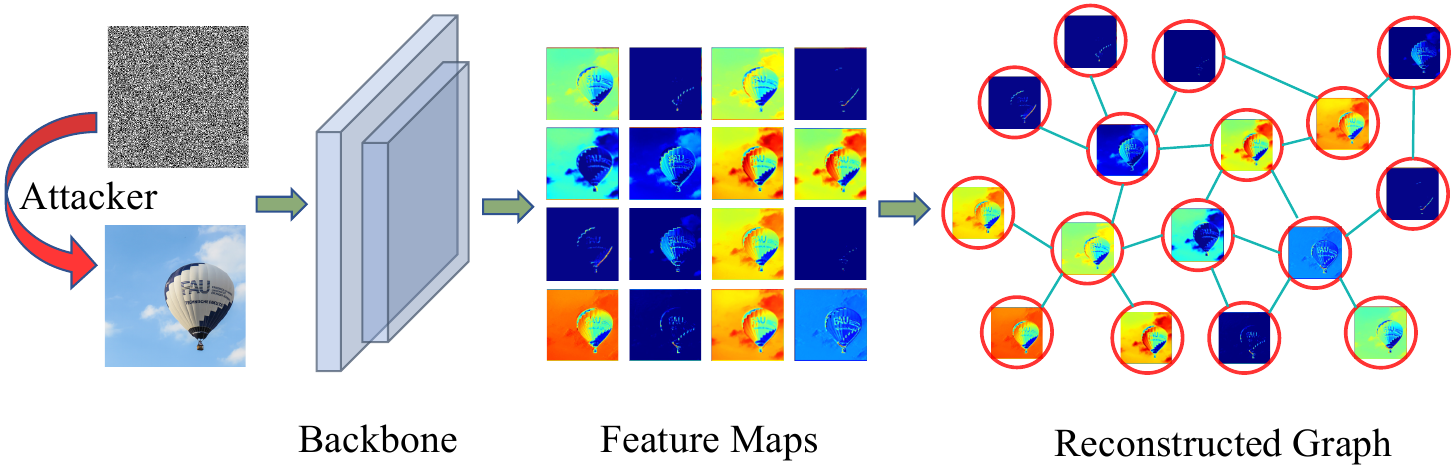}
    \end{center}
    \caption{The process of reconstructing the graph at the channel level by exploiting the similarity between feature maps.}
    \label{Graph_reconstructed_feature}
\end{figure}

In this paper, we introduce a plug-and-play module named Feature Map-based Reconstructed Graph Convolution (FMR-GC) to calibrate the contaminated features.
Specifically, we regard each feature map of the feature $\mathbf{X} \in \mathbb{R}^{H \times W \times C}$ extracted from a convolutional layer as a node. 
Then we construct a graph at the channel-wise level by selecting feature nodes with the highest similarity to better capture the inter-feature map correlations, as depicted in \Cref{Graph_reconstructed_feature}. 
Finally, we employ graph convolution~\citep{zhang2019graph} to aggregate neighborhood information for calibrating the contaminated features into normal state. 
Through extensive experiments, we verified that our FMR-GC outperforms state-of-the-art methods for defending the adversarial attack. 
Notably, FMR-GC can integrate seamlessly with existing AT techniques~\citep{PGD_AT, TRADES, AWP} for end-to-end training, thereby further augmenting model robustness. 
Meanwhile, our technique enhances model robustness without sacrificing the accuracy of clean samples.

\par 
In conclusion, our study has made the following contributions:
\begin{itemize}
    \item Unlike recent studies that that either deactivate or overactivate channel features, we introduce a novel approach of reconstructing the graph at the channel level and leveraging context information to calibrate contaminated features.
    \item We propose the Feature Map-based Reconstructed Graph Convolution (FMR-GC) module, a plug-and-play solution that reconstructs graphs within the channel dimensions and conducts graph convolution operations to calibrate contaminated features. 
    \item Our model achieves remarkable resilience performance with minimal additional computational cost. Moreover, it is designed to interact with advanced adversarial training methods, thereby further enhancing the model's robustness. 
\end{itemize}

\section{Related Work}
\subsection{Adversarial Attack}
Deep neural networks demonstrate a high vulnerability to meticulously crafted adversarial perturbations~\citep{wu2024towards, zakariyya2023towards}, which is attributed to their inherent linearity and limited flexibility~\citep{Taghanaki_2019_CVPR}. These perturbations accumulate through intermediate layers and eventually result in inaccurate predictions. 
\par 
Fast Gradient Sign Method (FGSM)~\citep{FGSM_attack} leverages gradient information to generate adversarial perturbations, drawing inspiration from the linear nature of neural networks.
Projected Gradient Descent (PGD)~\citep{PGD_attack} is an extension of FGSM, iteratively searching and updating perturbations in the disturbance space to create more powerful adversarial perturbations. 
Carlini-Wagner (CW)~\citep{CW_attack} generates adversarial perturbations by employing constrained optimization and compares three novel attack methods that utilize the $L_0$, $L_2$ and $L_{\infty}$ distance metrics.
AutoAttack (AA)~\citep{AA_attack} combines four distinct attack algorithms (\emph{i.e.,} APGD-CE, APGD-DLR, FAB~\citep{FAB_attack}, and Square Attack~\citep{Square_attack}), to create a parameter-free attack ensemble that evaluates adversarial robustness.

\subsection{Adversarial Defense}
Adversarial Training (AT) is widely recognized as the leading approach to enhance the robustness of deep neural network models. Its fundamental principle involves introducing imperceptibly small perturbations to the model's input and integrating the resulting adversarial examples into the training process. Various methods have been proposed to achieve this, such as AWP~\citep{AWP}, TRADES~\citep{TRADES}, FAT~\citep{FAT}, LBGAT~\citep{LBGAT}, PGD-AT~\citep{PGD_AT}, MART~\citep{MART}, SAT~\citep{SAT}. The procedure of AT can be expressed through a minimax optimization formulation:
\begin{equation}
    \label{AT_equation}
    \underset{\theta}{\min}\mathbb{E}_{(x,y)\sim \mathcal{D}}\underset{||\delta||_p \le \epsilon}{\max}CE(f_{\theta}(x+\delta),y),
\end{equation}
where $\mathcal{L}$ represents the loss function with respect to the model's parameter $\theta$, $(x,y)$ is a clean image-label pair sampled from the data distribution $\mathcal{D}$. Additionally, $\delta$ is a perturbation constrained by a maximum $p$-norm magnitude of~$\epsilon$.
\par
PGD-AT~\citep{PGD_AT} conducted a study on the occurrence of overfitting in robust AT and proposed employing a validation set protocol while performing model selection. 
TRADES~\citep{TRADES} improved the robustness of the model by introducing a trade-off loss term:
\begin{equation}
    \underset{\theta}{\min}\mathbb{E}_{(x,y)\sim \mathcal{D}}( CE(f_{\theta}(x),y)+\beta \cdot \underset{||\delta||_p \le \epsilon}{\max}KL(f_{\theta}(x),f_{\theta}(x+\delta))),
    \label{TRADES_equation}
\end{equation}  
where $KL(\cdot)$ means the KL divergence to constrain the distance between the classification accuracy distributions of clean samples $x$ and adversarial samples~$x+\delta$.
AWP~\citep{AWP} improve robustness by promoting the flatness of the weight loss landscape:
\begin{equation}
    \label{AWP_equation}
    \underset{\theta}{\min}\mathbb{E}_{(x,y)\sim \mathcal{D}}\underset{||\delta||_p \le \epsilon, \gamma \in \Gamma}{\max}(CE(f_{\theta+\gamma}(x+\delta),y)),
\end{equation}
where $\gamma$ denotes the injecting worst-case weights within the feasible region $\Gamma$ centered around $f_\theta$.
\par
While AT can significantly enhance model performance in the face of attacks, it fails to mitigate feature contamination instigated by adversarial perturbations during network inference~\citep{Xie_2019_CVPR}. 
This situation has stimulated several research initiatives aimed at guiding the calibration of the feature space, with the intent of learning more robust features. 
\citet{bai2020improving} explained the activation behavior of adversarial samples on features from the perspectives of magnitude and frequency, calibrating the feature space by suppressing non-robust activations. 
\citet{Ma_2021_ICCV} proposed training attention-guided generators and segmentation generators to learn robust medical vascular representations. 
\citet{zoran2020towards} used attention mechanisms to correct the attention logits in a single-channel feature map. 
\citet{kim2023feature} proposed reusing non-robust features to extract useful cues. 
\citet{FPCM} investigated how to configure the frequency characteristics of the feature space from a frequency perspective. 
These methods have successfully guided the model in learning more robust representations and in improving its defenses against adversarial samples. 
\par 
Despite these preceding methods, we propose a novel strategy for harmonizing feature maps. This process is realized by reconstructing graphs to capture correlations between feature maps and treating the features of neighboring nodes as embeddings within a latent space. 
Moreover, we employ graph convolution operations to calibrate features by leveraging neighbor contextual information. 
The method proposed in this study represents an orthogonal strategy to AT methods. When combined with advanced AT methods, this approach has the potential to further bolster the model's robustness.

\begin{figure*}[t]
        \begin{center}
        \includegraphics[width=\linewidth]{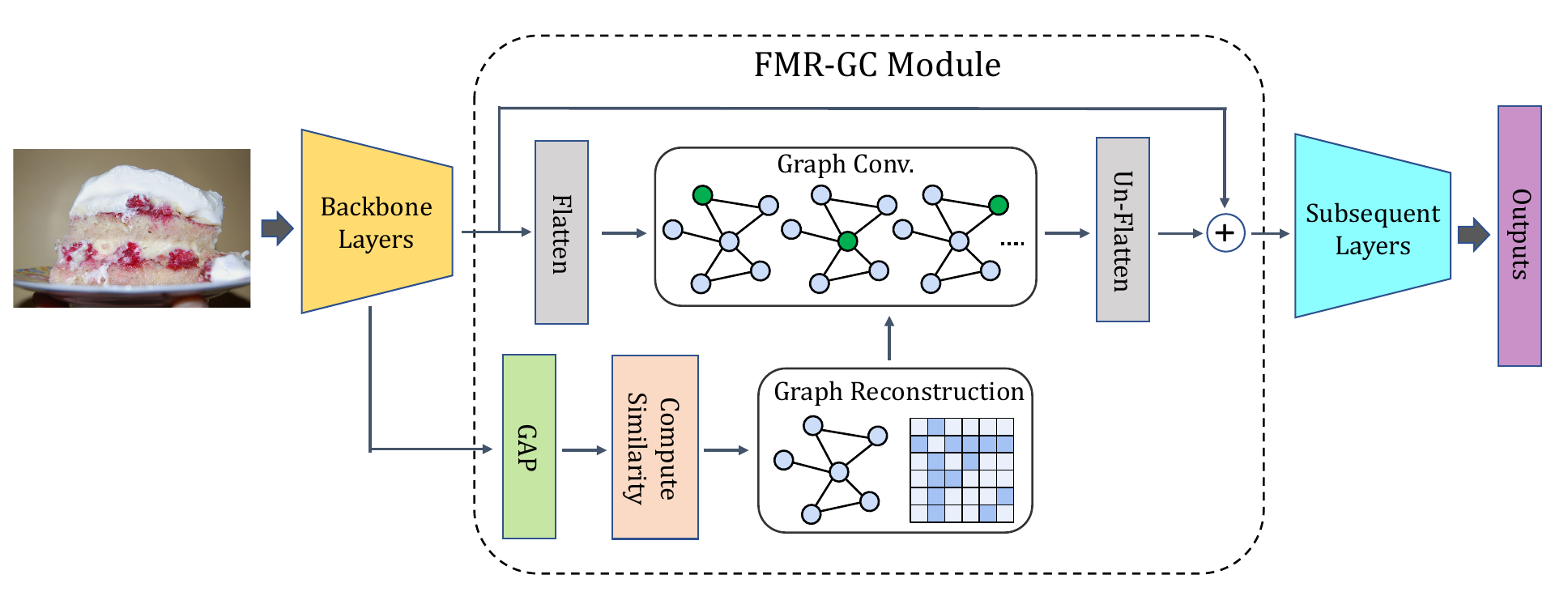}
        \end{center}
        \caption{The pipeline of our proposed model depicting the process. FMR-GC considers each feature map as a node and constructs a graph within the channel dimension using global average pooling and similarity computation. After reconstructing the graph, we perform graph convolution to calibrate contaminated features by leveraging contextual information.}
        \label{Model_overview}
\end{figure*}

\section{Method}
In this section, we first provide the definition of the problem tackled in this study. Subsequently, we explore the underlying motivation and the specific implementation of the graph reconstruction method employed in our approach. Lastly, we present the framework derived from our research and provide computational details of the proposed FMC-GC module.

\subsection{Problem definition}
The aims of this study is to propose a plug-and-play module, denoted as $\mathcal{G}(\theta_2)$, that harmonizes feature maps to calibrates the contaminated features. This module is designed for seamless integration into an existing convolutional neural network, denoted as $\phi(\theta_1)$. The primary objective of this integration is to enhance the model's resilience against adversarial attacks through adversarial training. 
This training approach is formulated as the following optimization problem:
\begin{equation}
    \underset{\theta_1,\theta_2}{\min}\mathbb{E}_{(x,y)\sim \mathcal{D}}\underset{||\delta||_p \le \epsilon}{\max}\mathcal{L}(\phi(\theta_1),\mathcal{G}(\theta_2);x+\delta,y),
    \label{AT}
\end{equation}
where $\mathcal{L}$ represents the loss function with respect to the model's parameters $\theta_1$ and $\theta_2$. $(x,y)$ denotes a clean image-label pair sampled from the data distribution $\mathcal{D}$. The perturbation $\delta$ is constrained by a maximum $l_p$-norm magnitude of $\epsilon$.

\subsection{Constructing Inter-Feature Map Correlations}
\label{Feature_Correlation}
Given the feature $\mathbf{X} \in \mathbb{R}^{H \times W \times C}$ extracted from a convolutional layer, we treat each feature map $x_i$ as a node, resulting in a set of nodes $\mathcal{V}=\{v_1,v_2,\dots, v_C\}$, where each node $v_i$ corresponds to the feature $x_i \in \mathbb{R}^{H \times W}$ of the channel $i$. 
To capture the relationships between nodes and reconstruct them as a graph, we apply the global average pooling (GAP) on each feature channel $\{\bm{x}_1,\bm{x}_2,\dots,\bm{x}_C\}$. 
Then, we have the mean-pooled feature vectors $\{\bar{\bm{x}}_1,\bar{\bm{x}}_2,\dots,\bar{\bm{x}}_C\}$.
This step serves to get the first-order statistic of the feature maps, which is beneficial for helping mitigate the impact of perturbations on the computation of reconstruction errors~\citep{chen2021adversarial}. Additionally, global average pooling contributes to reducing computational costs in the subsequent stages.
Subsequently, we construct a similarity matrix $S$ based their Euclidean distance:
\begin{equation}
    S_{i,j} = \exp(-\frac{\Vert\bar{\bm{x}_i}-\bar{\bm{x}_j}\Vert^2}{\sigma^2}),
\label{S_matrix}
\end{equation}
where $S_{i,j}$ represents the similarity score between two feature maps $\bar{\bm{x}_i}$ and $\bar{\bm{x}_j}$, and $\sigma$ is a parameter that controls the decay rate of the similarity with the distance.
To ensure that feature maps are not compared with themselves, we set the diagonal entries of $S$ to negative infinity, \textit{i.e.}, $S_{i, i} = -\infty$.
\par
In detail, for each node corresponding to $\bar{\bm{x}_i}$, we examine the $i$-th row of $S$, which contains the similarity scores between $\bar{\bm{x}_i}$ and all other feature maps. We then choose the top-$k$ nodes with the highest similarity scores and connect $\bar{\bm{x}_i}$ to these nodes, creating an edge in $E$. By doing so, we reconstruct $G=(V, E)$ to better reflect the inter-feature map correlations.
With the graph $G$ and the flattened feature map vectors $\bm{f}_i = vec(\bm{x}_i)$ for $i = 1,2,...,n$, we can represent the graph signal matrix $F$ as:
\begin{equation}
    \label{graph_signal}
    F = [\bm{f}_1,\bm{f}_2,...,\bm{f}_c]^\top,
\end{equation}
where each $\bm{f}_i$ corresponds to the flattened feature map of node $i$.

\subsection{Graph Convolution Processing}
\label{GCN_design}
In the previous section, we discussed a channel-level graph reconstruction process applied to the feature map, resulting in the generation of a graph $G$. 
Modeling a graph at the channel level offers two notable advantages. First, the graph connecting feature maps capture contextual information and spatial associations, allowing for improved feature extraction through neighborhood information aggregation~\citep{li2020multi, NIPS2014_7810ccd4}. 
This is achieved by employing the neighborhood aggregation function, denoted as:
\begin{equation}
    \mathcal{G}(\theta_2) = \sigma(Agg_{\theta_2}({F_v|v \in \mathbb{N}(v)})),
    \label{Neibor_Agg}
\end{equation}
where $Agg_{\theta_2}$ denotes the neighborhood aggregation function, $\mathbb{N}(v)$ represents neighbors of the node $v$.
\par 
In the second place, by incorporating neighborhood information of nodes to defense attacks, one can potentially mitigate the contaminated feature activations caused by adversarial perturbations.
To achieve this, we propose the Feature Map-based Reconstructed Graph Convolution (FMR-GC) method, which is outlined in \Cref{Model_overview}.
\par
Our proposed graph-based module takes advantage of the neighborhood aggregation characteristic of Graph Convolutional Networks (GCNs) to treat the input flattened feature maps $F$ (as defined in Eq. \ref{graph_signal}), alongside the graph $G = (V, E)$ that was constructed using inter-feature map information (Section \ref{Feature_Correlation}). The GCN convolution operation, inclusive of residual connections, is articulated as follows:
\begin{equation}
\mathcal{G}(\theta_2) = \sigma(\tilde{D}^{-\frac{1}{2}}\tilde{A}\tilde{D}^{-\frac{1}{2}}X\theta_2) + F
\label{GCN_Eq}
\end{equation}
Here, $\tilde{A}$ denotes the adjacency matrix of the graph $G$ with self-connections, $\tilde{D}$ is the corresponding diagonal degree matrix, $F$ is the input feature, and $\theta_2$ is the weight matrix. After applying the GCN convolution to the input graph signal matrix, we reshape the output to the original dimension to obtain the calibrated feature maps.
\par
FMR-GC functions as a plug-and-play module that can be easily incorporated into existing CNN architectures $\phi(\theta_1)$ and applied to several convolutional layers within CNNs.

\section{Experiments}
In this section, we present the experimental setup for this study, including the training details and the testing methods employed. Subsequently, a series of experiments are conducted to assess the performance and scalability of the proposed model. Additionally, we explore the influence of different representation features and sparsity on the reconstructed graph and the robust performance through an in-depth analysis. 
\subsection{Experimental Setup}
\textbf{Implementation Details:}
We evaluate the robustness of our proposed approach through experiments conducted on CIFAR-10~\citep{CIFAR}, CIFAR-100~\citep{CIFAR}, and Tiny ImageNet~\citep{deng2009ImageNet} datasets. 
Baseline models used for performance comparison include WRN34-10~\citep{zagoruyko2016wide}, ResNet-18~\citep{He_2016_CVPR}, VGG16~\citep{VGG}, WRN32-10~\citep{zagoruyko2016wide}, and Inc-V3~\citep{Inc-V3}.
\par 
We integrate our model with three AT methods: PGD-AT~\citep{PGD_AT}, TRADES~\citep{TRADES}, and AWP~\citep{AWP} to validate its wide applicability. The combinations were labeled as FMR-GC-AT, FMR-GC-TRADES, and FMR-GC-AWP.
For a comparative analysis with other models (see in \Cref{compare_robust}), all models were trained using the same hyperparameters and training details outlined in their respective original papers~\citep{PGD_AT, TRADES, AWP}. Additionally, the parameters for the two feature calibration methods, FSR~\citep{kim2023feature} and FPCM~\citep{FPCM}, are set to align with those of PGD-AT~\citep{PGD_AT}.
For additional performance assessment, models undergo a training of 100 epochs, initiating with a learning rate of 0.1. This rate is reduced by a factor of 0.1 at the 90th and 95th epochs. Optimization was performed using the SGD optimizer, with a momentum of 0.9 and a weight decay factor of 5e-4.
During network inference, the reconstructed graph $G$ is generally considered to be undirected with $k=5$. 
Experiments were executed on a system equipped with two NVIDIA RTX-A4000 GPUs.
\par \textbf{Evaluation Settings:}
The model's robustness is evaluated using a variety of attack methods, including FGSM~\citep{FGSM_attack}, PGD~\citep{PGD_attack}, C\&W~\citep{CW_attack}, and AutoAttack~\citep{AA_attack}. Notably, AutoAttack constitutes APGD-DLR~\citep{AA_attack}, APGD-CE~\citep{AA_attack}, FAB~\citep{FAB_attack}, and Square~\citep{Square_attack}. Unless otherwise specified, these attacks are performed under the $L_\infty$ norm with $\epsilon$ set to 8. It is important to mention that ``Clean" refers to the accuracy of the original test samples.

\begin{table*}[t]
	\begin{center}
		\caption{Test robustness~(\%) on CIFAR-10 using WRN34-10. The \textbf{number} in bold indicates the best accuracy.}
		\label{SOTA_Comparison}
		\resizebox{0.98\textwidth}{!}
		{
			\begin{tabular}{clccccccc}
				\toprule
				Dataset&Method& Publish& Clean& PGD-10& PGD-20& PGD-50& C\&W& AA\\ \midrule
				\multicolumn{1}{c}{\multirow{11}{*}{CIFAR-10}}&PGD-AT~\cite{PGD_AT}             &ICML-20& 85.17& 56.07& 55.08& 54.88& 53.91& 51.69\\ 
				&FPCM~\citep{FPCM}  &ICCV-23& 85.67& 56.72& 55.84& 55.40& 54.61& 52.71\\ 
				&MART~\citep{MART}                 &ICLR-19& 84.17& 58.98& 58.56& 58.06& 54.58& 51.10\\ 
				&TRADES~\citep{TRADES}             &ICML-19& 85.72& 56.75& 56.10& 55.90& 53.87& 53.40\\ 
				&LBGAT~\citep{LBGAT}               &ICCV-21& \textbf{88.22}& 56.25& 54.66& 54.30& 54.29& 52.23\\
				&FAT~\citep{FAT}                  &ICML-20& 87.97 & 50.31 & 49.86 & 48.79 & 48.65 & 47.48 \\
				&FSR~\citep{kim2023feature}        &CVPR-23& 83.96 & 55.94 &55.11 & 54.71 &54.46 & 52.35 \\
                    &GAIRAT~\citep{GAIRAT}            &ICLR-20& 86.30 & 59.64 & 58.91 & 58.74 & 45.57 & 40.30 \\
				&AWP~\citep{AWP}                   &NeurIPS-20& 85.57& 58.92& 58.13& 57.92& 56.03& 53.90\\
				\cmidrule{2-9}
				&FMR-GC-AT     &\multicolumn{1}{c}{\multirow{3}{*}{Ours}}&86.80&58.97&58.33&57.54&55.75&54.64\\
				&FMR-GC-TRADES     &&85.58&59.54&58.97&58.71&54.88&54.25\\
				&FMR-GC-AWP    &&85.62&\textbf{61.05}&\textbf{59.97}&\textbf{59.41}&\textbf{57.03}&\textbf{56.37}
				\\ \midrule 
				\multicolumn{1}{c}{\multirow{8}{*}{CIFAR-100}}&PGD-AT~\cite{PGD_AT}   &ICML-20& 60.89& 32.19& 31.69& 31.45& 30.10& 27.86\\ 
				&TRADES~\citep{TRADES}   &ICML-19& 58.61& 29.20& 28.66& 28.56& 27.05& 25.94\\
				&LBGAT~\citep{LBGAT}     &ICCV-21& 60.64& 35.13& 34.75& 34.62& 30.65& 29.33\\
				&SAT~\citep{SAT}         &AISec-21& \textbf{62.82}& 28.10& 27.17& 26.76& 27.32& 24.57\\ 
				&AWP~\citep{AWP}         &NeurIPS-20& 60.38& 34.13& 33.86& 33.65& 31.12& 28.86\\ 
				\cmidrule{2-9}
				&FMR-GC-AT     &\multicolumn{1}{c}{\multirow{3}{*}{Ours}}&61.88&34.83&34.24&33.76&31.59&29.80\\
				&FMR-GC-TRADES     &&59.88&32.87&32.48&32.24&28.97&28.13\\
				&FMR-GC-AWP    &&60.65&\textbf{36.84}&\textbf{35.93}&\textbf{35.82}&\textbf{32.02}&\textbf{31.04}
				\\ \bottomrule
			\end{tabular}
		}
	\end{center}
\end{table*}

\subsection{Comparison with Other Methods}
\label{compare_robust}
\subsubsection{Comparisons on CIFAR-10 and CIFAR-100}
In this part, we compared the performance of our proposed method with other existing approaches. We utilize WRN34-10 as the target network and incorporate a single FMR-GC block after the initial convolutional layer, which converts the image into feature maps. The experimental results on CIFAR-10 and CIFAR-100 are reported in \Cref{SOTA_Comparison}. We compare our method against the following baseline approaches: 1)~PGD-AT~\citep{PGD_AT}, 2)~FPCM~\citep{FPCM}, 3)~FSR~\citep{kim2023feature} 4)~MART~\citep{MART}, 5)~GAIRAT~\citep{GAIRAT}, 6)~FAT~\citep{FAT}, 7)~TRADES~\citep{TRADES}, 8)~LBGAT~\citep{LBGAT} and 9)~AWP~\citep{AWP}. 
\par 
Our analysis from \Cref{SOTA_Comparison} is outlined as follows:
\par 1) \textbf{AT methods equipped with FMR-GC can improve adversarial robustness.} 
The integration of the FMR-GC module into AT strategies consistently enhances their performance across various attack scenarios. For instance, on the CIFAR-10 dataset, the incorporation of our FMR-GC module into PGD-AT yielded performance enhancements of 2.90\%, 1.84\%, and 2.95\% against PGD-10, C\&W, and AA attacks, respectively. These results underscore the adaptability of our approach in improving the robustness of AT methods efficiently.
\par 2) \textbf{AT models equipped with FMR-GC exhibit superior performance against all types of attacks.} 
The three proposed models showcased exceptional performance across most attack scenarios, even when integrating FMR-GC into the basic PGD-AT strategy.
The utilization of the AWP strategy equipped with the FMR-GC module resulted in achieving state-of-the-art performance. In comparison to suboptimal methods, FMR-GC-AWP demonstrated accuracy improvements of 1.41\%, 1.01\%, and 2.97\% on PGD-10, C\&W, and AA attacks on the CIFAR-10 dataset, respectively.
\par 3) \textbf{FMR-GC improves robustness without sacrificing accuracy on clean samples.} 
Our model shows a better trade-off between maintaining accuracy on clean samples and enhancing robustness compared to baseline methods. 
By incorporating the FMR-GC module into the PGD-AT, TRADES, and AWP strategies, our model consistently improves the accuracy of the clean samples and enhances adversarial robustness. 
Notably, when trained with PGD-AT, our FMR-GC-AT model exhibited remarkable improvements of 1.63\% and 0.99\% on the CIFAR-10 and CIFAR-100 datasets, respectively, compared to the base method.

\begin{table*}[t]
	\begin{center}
		\caption{Test robustness~(\%)  on Tiny-ImageNet using ResNet18. The \textbf{number} in bold indicates the best accuracy.}
		\label{ImageNet-Robust}
		\resizebox{0.75\linewidth}{!}
		{
			\begin{tabular}{l|cccccc}
				\hline
				Method           & Clean          & PGD-10         & PGD-20         & PGD-50         & C\&W           & AA             \\ \hline 
			PGD-AT~\cite{PGD_AT} &44.15&21.45&21.08&20.91&18.74&16.13\\
                PGD-AT + FMR-GC   &\textbf{46.86}&\textbf{24.55}&\textbf{23.94}&\textbf{23.62}&\textbf{20.84}&\textbf{18.33}\\ 
                \midrule
                \midrule
                TRADES~\cite{TRADES} &45.13&20.37&20.08&19.95&16.17&14.42\\
                TRADES + FMR-GC  &\textbf{46.20}&\textbf{24.97}&\textbf{24.62}&\textbf{24.05}&\textbf{19.17}&\textbf{18.04}\\
			\midrule
                \midrule
			AWP~\cite{AWP}    &44.86&22.03&21.72&21.50&19.08&16.97\\
                AWP + FMR-GC      &\textbf{45.92}&\textbf{26.15}&\textbf{25.73}&\textbf{25.34}&\textbf{21.81}&\textbf{19.93}\\
			\hline
			\end{tabular}
		}
	\end{center}
\end{table*}

\begin{table*}[t]
    \caption{Transfer attack accuracy~(\%) in the single-model transfer scenario. The \textbf{number} in bold indicates the best accuracy.}
    \label{trans_attack}
    \begin{center}
    \resizebox{0.85\textwidth}{!}
    {
    \begin{tabular}{c|ccc|ccc}
        \toprule
        \multirow{3}*{Attack ($\epsilon=8$)} & \multicolumn{6}{c}{Performance w/o and w/ FMR-GC}\\
        \cmidrule{2-7}
         & \multicolumn{3}{c|}{Source: WRN34-10} & \multicolumn{3}{c}{Source: ResNet18} \\
        & $\Rightarrow$ ResNet18 &$\Rightarrow$ VGG16 & $\Rightarrow$ Inc-v3&$ \Rightarrow$ WRN34-10 &$\Rightarrow$ VGG16 & $\Rightarrow$ Inc-v3\\ \midrule
        Natural &$85.32/\textbf{86.80}$&$85.32/\textbf{86.80}$&$85.32/\textbf{86.80}$&$80.18/\textbf{81.35}$&$80.18/\textbf{81.35}$&$80.18/\textbf{81.35}$\\
        FGSM    &$81.00/\textbf{83.57}$&$81.52/\textbf{83.66}$&$80.95/\textbf{82.31}$&$77.14/\textbf{77.73}$&$77.39/\textbf{78.40}$&$77.12/\textbf{77.90}$\\
        PGD-10  &$66.11/\textbf{67.36}$&$69.51/\textbf{70.43}$&$65.48/\textbf{67.40}$&$61.34/\textbf{62.69}$&$62.10/\textbf{63.70}$&$60.66/\textbf{61.48}$\\
        PGD-20  &$65.16/\textbf{66.57}$&$69.23/\textbf{70.12}$&$65.40/\textbf{67.02}$&$60.91/\textbf{62.01}$&$62.11/\textbf{63.30}$&$60.44/\textbf{61.09}$\\
        AA      &$70.36/\textbf{72.13}$&$76.97/\textbf{78.55}$&$69.40/\textbf{70.11}$&$64.71/\textbf{65.42}$&$69.73/\textbf{70.41}$&$63.68/\textbf{64.90}$\\
        \bottomrule
    \end{tabular}
    }
    \end{center}
\end{table*}

\subsubsection{Comparisons on Tiny-ImageNet}
In this part, we evaluate the model's generalization capability on a large-scale dataset by integrating the FMR-GC module into ResNet-18 and conducting evaluations on the Tiny-ImageNet dataset. 
Tiny-ImageNet presents an increased resolution and a larger set of classification categories. The results of these experiments are presented in \Cref{ImageNet-Robust}. Models equipped with the FMR-GC module demonstrated notable improvements in robust accuracy over the baseline while maintaining high accuracy on clean samples. Specifically, FMR-GC-TRADES exhibited significant performance gains over the FMR-GC module, achieving accuracy improvements of 1.07\% and 4.60\% for Clean and PGD-10, respectively. Moreover, the AWP-FMR-GC model emerged as the top-performing model, surpassing the AWP method by 4.12\% and 2.96\% on PGD-10 and AA metrics, respectively. These findings suggest that our proposed module seamlessly integrates with various AT techniques, notably enhancing their effectiveness against adversarial attacks.

\begin{table*}[t]
    \caption{Test robustness~(\%) on CIFAR-10 using ResNet18 under adversarial training. The \textbf{number} in bold indicates the best accuracy.}
    \label{AT_train}
    \begin{center}
    \resizebox{0.85\textwidth}{!}
    {
    \begin{tabular}{cccccccccc}
        \toprule
         \multicolumn{3}{c}{Denoising Block Position} & \multirow{2}*{Natural}& \multicolumn{6}{c}{Accuracy under White-box Attack ($\epsilon= 8$)}  \\ 
        \cmidrule(r){1-3} \cmidrule(r){5-10}
        Conv.1 & Conv.2 & Conv.3 & &FGSM & PGD-10 &PGD-20 & PGD-50 & C\&W &AA \\
        \midrule
        &&&80.18&74.30&51.92&51.35&50.95&48.90&47.10\\
        \midrule
        \checkmark&&                    &81.35&76.52&54.91&53.89&52.98&50.86&49.84\\
        &\checkmark&                    &80.94&75.80&53.75&53.20&52.69&50.16&48.64\\
        &&\checkmark                    &80.90&75.56&53.43&52.98&52.61&50.04&48.53\\
        \checkmark&\checkmark&          &81.56&76.94&55.03&53.97&53.07&51.72&50.72\\
        \checkmark&\checkmark&\checkmark&\textbf{81.77}&\textbf{77.11}&\textbf{55.17}&\textbf{54.12}&\textbf{53.26}&\textbf{52.83}&\textbf{51.01}\\
        \bottomrule
    \end{tabular}
    }
    \end{center}
\end{table*}

\subsection{Robustness to Transfer Attacks}
In this part, we investigated the ability of the FMR-GC equipped model to defend against transfer attacks. 
In scenarios where attackers lack access to the network's architecture details, the feasibility of white-box threats diminishes, leading attackers to employ alternative models for crafting transfer attacks.
Evaluating the effectiveness of transfer attacks provides evidence that enhanced robustness is not the result of gradient masking.
To evaluate the performance of the WRN34-10 and ResNet18 models integrated with the FMR-GC module against transfer attacks, adversarial samples were generated using ResNet18, Inc-v3, VGG16, and WRN34-10 models. The results, illustrating the model response to transfer attacks, are presented in \Cref{trans_attack}, leading to the following conclusions:
\par 1) \textbf{FMR-GC's effective defense against various transfer attacks.} 
Integrating the FMR-GC module significantly improved model performance in defending different transfer attacks compared to the baseline.
Specifically, for the WRN34-10 model equipped with a single FMR-GC module at Conv.1, utilizing the Inc-v3 model as the source model to generate adversarial samples, the defense success rate witnessed an improvement of 1.36\%, 1.92\%, and 0.71\% on FGSM, PGD-10, and Autoattack, respectively, as compared to the original WRN34-10 model.
\par 2) \textbf{FMR-GC's effective defense against transfer attacks from diverse source models.} 
Incorporating the FMR-GC module into our model proved effective in defending against transfer attacks crafting from various source models. This highlights the model's robustness, extending beyond specific attack scenarios and demonstrating reliability against attacks from diverse model architectures. For instance, when the ResNet18 model is equipped with the FMR-GC module and subjected to PGD-10 attacks generated by WRN34-10, VGG16, and Inc-V3 models, the defense success rate increases by 1.35\%, 1.60\%, and 0.88\%, respectively, compared to the original model.

\par 3) \textbf{Universal defense provided by FMR-GC across multiple models.} 
Equipping FMR-GC with various models significantly enhances their effectiveness in defending against transfer attacks. Specifically, when faced with the Autoattack generated by Inc-V3, both the WRN34-10 and ResNet18 models equipped with the FMR-GC module exhibited notable improvements in success rates. The WRN34-10 model demonstrated a success rate increase of 0.71\%, while the ResNet18 model showed an even greater increase of 1.22\% compared to their counterparts without the FMR-GC module.

\subsection{Exploring Performance with Varying Block Positions}
In this part, we investigated the impact of equipping FMR-GC at different positions during model inference. 
In \Cref{AT_train}, we analyzed the performance of equipping FMR-GC at different positions within the convolutional layers, while conducting adversarial training. Our investigations have led to the following conclusions:
\par 1) \textbf{Significant performance enhancement by early integration of FMR-GC in the inference process.} 
Integrating FMR-GC at the initial stages of model inference brings about notable performance improvements. 
When computational resources are constrained, incorporating a single FMR-GC module reduces the computational burden. Inserting the module into the early stages of model inference results in a more substantial enhancement in model performance. This improvement can be attributed to the larger dimensions and lower abstraction levels of features in the reconstructed graph nodes during the early stages of inference. For example, equipping FMR-GC at the Conv.1 layer yielded a 2.99\% improvement in PGD-10 and a 2.74\% improvement in Autoattack, compared to equipping it at the Conv.2 and Conv.3 layers.
\par 2) \textbf{Enhanced model performance through multiple FMR-GC module deployments.} Our research revealed that deploying multiple FMR-GC modules concurrently at varied inference positions significantly enhances model robustness. Furthermore, there exists a positive correlation between the number of deployments and performance, proving advantageous in scenarios with ample computational resources. For instance, equipping three modules at Conv.1, Conv.2 and Conv.3, rather than only equipping at the Conv.1 layer, can increase the success rate of PGD-10 and C\&W attacks by 1.97\% and 1.17\% respectively.

\begin{table*}[t]
    \caption{Test robustness~(\%) on CIFAR-10 using WRN32-10 under simple training. The \textbf{number} in bold indicates the best accuracy.}
    \label{simple_train}
    \begin{center}
    \resizebox{0.85\textwidth}{!}
    {
    \begin{tabular}{cccccccccc}
        \toprule
         \multicolumn{3}{c}{Denoising Block Position} & \multirow{2}*{Natural}& \multicolumn{6}{c}{Accuracy under White-box Attack ($\epsilon= 8$)}  \\ 
        \cmidrule(r){1-3} \cmidrule(r){5-10}
        Conv.1 & Conv.2 & Conv.3 & &FGSM & PGD-10 &PGD-20 & PGD-50 & C\&W &AA \\
        \midrule
        &&&95.88&53.25&16.79&16.72&16.61&14.08&13.73\\
        \midrule
        \checkmark&&&95.76&55.84&23.95&23.87&23.76&21.08&24.77\\
        &\checkmark&&95.85&54.80&18.70&18.61&18.49&16.70&18.21\\
        &&\checkmark&96.03&53.55&17.29&17.21&16.97&14.63&16.89\\
        \checkmark&\checkmark&&95.91&59.64&27.36&27.32&27.04&25.38&29.24\\
        \checkmark&\checkmark&\checkmark&\textbf{96.13}&\textbf{59.84}&\textbf{28.33}&\textbf{28.19}&\textbf{27.95}&\textbf{24.72}&\textbf{30.19}\\
        \bottomrule
    \end{tabular}
    }
    \end{center}
\end{table*}

\subsection{Performance without Adversarial Training}
In this part, we evaluated the performance of our proposed model under simple training (ST). AT incurs significant computation costs and may adversely affect the accuracy of clean samples, making simple training the only feasible option in certain scenarios. Simple training entails operating without artificially generated adversarial samples as inputs. In \Cref{simple_train}, we evaluated the performance of both the original model and the model equipped with FMR-GC modules at different position under simple training. Our findings are as follows:
\par 1) \textbf{FMR-GC enhances the robustness without sacrificing its accuracy under simple training.}
By incorporating the FMR-GC module into the model and employing computationally efficient simple training, we observed a substantial improvement in the model's resilience against various attack scenarios. For example, when introducing a single FMR-GC module after Conv.1, he model exhibited significant performance gains of 6.16\% against PGD-10, 7.00\% against C\&W, and 11.04\% against Autoattack attacks compared to the baseline model.
\par 2) \textbf{The performance of the model correlates with the placement and quantity of FMR-GC modules.} 
The model's performance demonstrates a strong correlation with the positioning and number of FMR-GC modules. 
Specifically, integrating the FMR-GC module early in the model's inference process significantly enhanced performance.
Specifically, integrating the FMR-GC module in the early stages of the model's inference process led to substantial performance improvements. Additionally, increasing the number of FMR-GC modules further enhanced the model's performance. A comparison between the insertion of a single FMR-GC module and multiple modules placed at Conv.1, Conv.2, and Conv.3 revealed notable performance improvements. Importantly, inserting the FMR-GC module at Conv.1 had a more pronounced effect compared to Conv.2 and Conv.3, resulting in a larger performance boost.

\subsection{Analysis The Influence of Reconstructed Sparse Graphs}
\begin{figure}[t]
    \begin{center}
    \includegraphics[width=\linewidth]{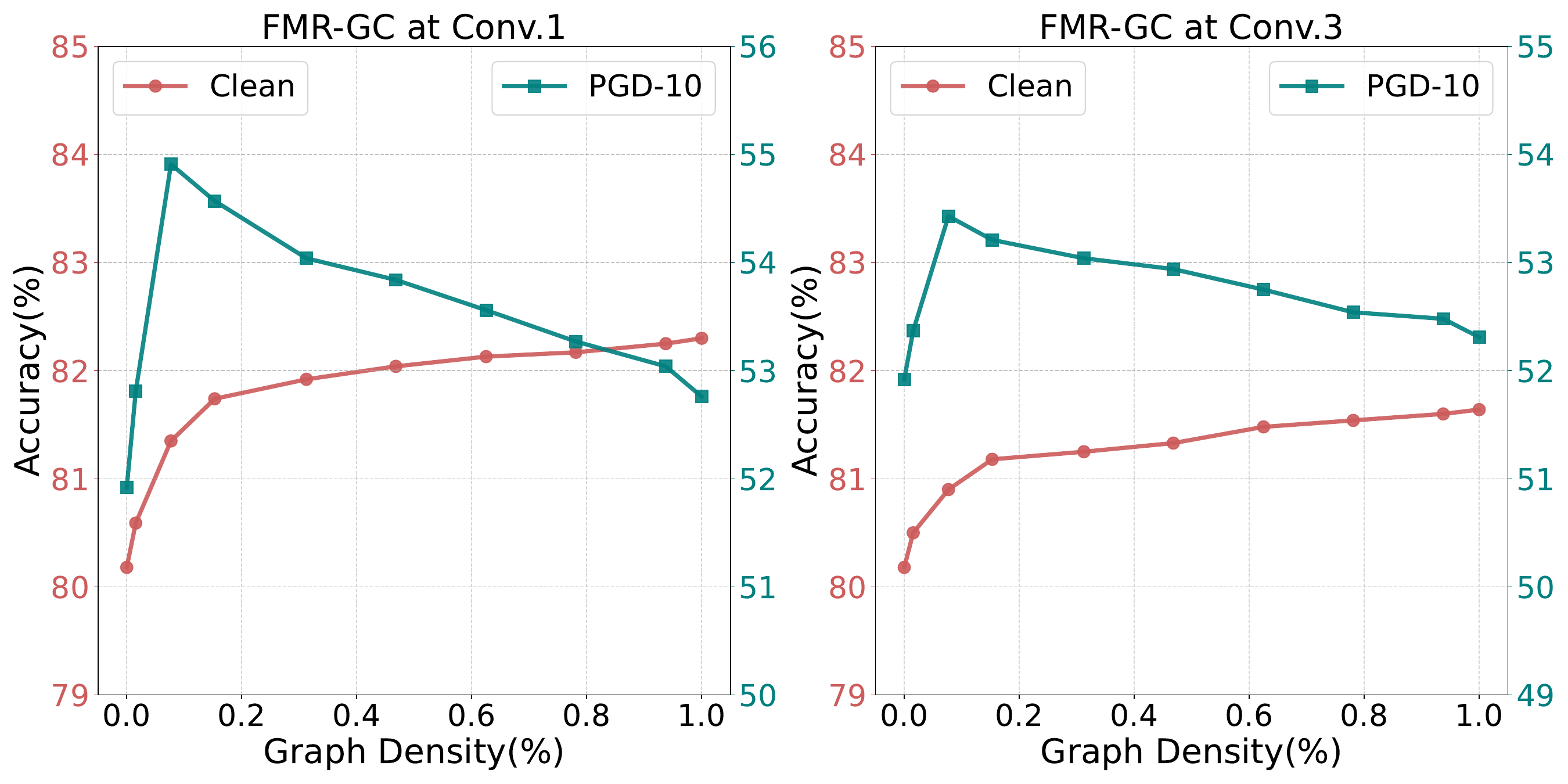}
    \end{center}
    \caption{Analysis of the influence of reconstruction graph sparsity on model performance on CIFAR-10 using ResNet18. The $x$-axis represents the graph sparsity, while the $y$-axis represents the accuracy~(\%).} 
    \label{graph_density:figure}
\end{figure}

In this part, we investigated how the sparsity of the reconstruction graph impacts the proposed FMR-GC module.
FMR-GC employs a unique hyperparameter $k$ to represent the selected top-$k$ most similar nodes in the reconstructed graph $G=(V, E)$. This parameter $k$ determines the average degree during the feature map reconstruction, thus indicating the graph's density levels. A higher value for $k$ produces a denser graph, and a lower value results in a sparser one. We express the graph's density as $\frac{k}{c}$, where $c$ is the number of channels in the FMR-GC's input features. Based on \Cref{graph_density:figure}, we draw the following conclusions:
\par 1) \textbf{Inference provided by sparse reconstruction enhances robustness in FMR-GC-equipped models.}
As the density of the reconstruction graph increases, there is a noticeable decrease in the model's robustness. This decrease is attributed to adversarial perturbations within the feature maps, leading to incorrect connections within the dense graph. 
During the model inference process, these inaccurate details might inadvertently be used within the graph convolution, leading to a reduction in calibration effectiveness. Notably, extreme sparsity in the graph results in a slight performance decline as it fails to accurately model interconnections among nodes, negatively impacting FMR-GC performance.
\par 2) \textbf{The density of the reconstruction graph directly influences the accuracy of clean samples.}
An interesting discovery reveals that while the FMR-GC module demonstrates superior robustness with a sparse reconstruction graph, the model's feature extraction capability for clean samples improves as the density increases. This enhancement is due to the reduced likelihood of erroneous connections in reconstructing the clean graph. The dense graph's interconnected relationships improve the understanding of correlations and contextual information among nodes, consequently enhancing performance. To strike a balance between robustness and accuracy for clean samples, we selected a hyperparameter of $k=5$ for the FMR-GC module.

\begin{table}[t]
    \caption{Test robustness~(\%) with various graph node feature representations on CIFAR10 using ResNet-18 equipped FMR-GC at Conv.1 . The \textbf{number} in bold indicates the best accuracy.}
    \label{S:tb_F}
    \begin{center}
    \resizebox{\columnwidth}{!}
    {
    \begin{tabular}{c|cccc}
        \toprule
        Method & Clean & PGD-10 & C\&W & AA\\ 
        \midrule
        Original-ResNet18&80.18&51.92&48.90&47.10\\
        \midrule
        Global-Avg-Pool&\textbf{81.35}&\textbf{54.91}&\textbf{50.86}&\textbf{49.84}\\
        W/o Pool&80.78&54.19&49.97&48.70\\
        Avg-Pool(kernel-size=4)&81.29&54.30&50.14&49.52\\
        Avg-Pool(kernel-size=8)&81.05&54.58&50.41&49.62\\
        \bottomrule
    \end{tabular}
    }
    \end{center}
\end{table}

\subsection{Analyzing Graph Reconstruction with Different Node Feature Representations}
In this part, we analyzed the effectiveness of utilizing different node feature representations when reconstructing graphs.
Each feature map in FMR-GC corresponds to a node, and the node feature representation is crucial for the reconstruction of the graph. We utilized several different techniques to extract node information before the reconstruction of the graph, namely: 1) global pooling-applied features, 2) original feature maps, and 3) average pooling-applied features with a kernel size of 4 and 8.
\Cref{S:tb_F} represents the performance of the model assessed with diverse node feature representation methods. Upon observation, the following conclusions were drawn: 
\par 1) \textbf{Reconstruction features in FMR-GC demonstrate varying performance levels, all outperform the original model.}
The model equipped with FMR-GC surpasses the performance of the original feature representation when multiple graph reconstruction features are utilized. For example, even the least effective method, W/o Pool, in FMR-GC demonstrates improvements of 0.60\%, 2.27\%, 1.07\%, and 1.60\% on Clean, PGD-10, C\&W, and AutoAttack, respectively.
This finding highlights the superiority of our proposed module and validates the correctness of its underlying motivation.
\par 2) \textbf{The performance of FMR-GC is directly influenced by the range of the pooling kernel.}
We find that global pooling surpasses average pooling in terms of robustness. Additionally, increasing the kernel size of average pooling leads to improved robustness performance. Conversely, the feature representation method without pooling exhibits comparatively inferior performance. This discrepancy arises from the fact that a larger pooling range has a diminished effect on local anomalous, facilitating the better integration of feature maps affected by adversarial perturbations and yielding a more stable graph representation.

\subsection{Performance under Diverse Amplitude Attacks}
In this part, we evaluated the performance of the proposed model against diverse amplitude attacks. These attacks involve manipulating the amplitudes of the input signals to assess the model's robustness against different levels of perturbations. We conducted PGD-10 and PGD-100 attacks on the WRN34-10 model, which was equipped with an FMR-GC module at Conv1. Perturbations of varying magnitudes, denoted as $\epsilon$, were applied during these attacks. From the results in \Cref{epison_ana}, we can have the following observations:

\begin{figure}[t]
    \begin{center}
    \includegraphics[width=\linewidth]{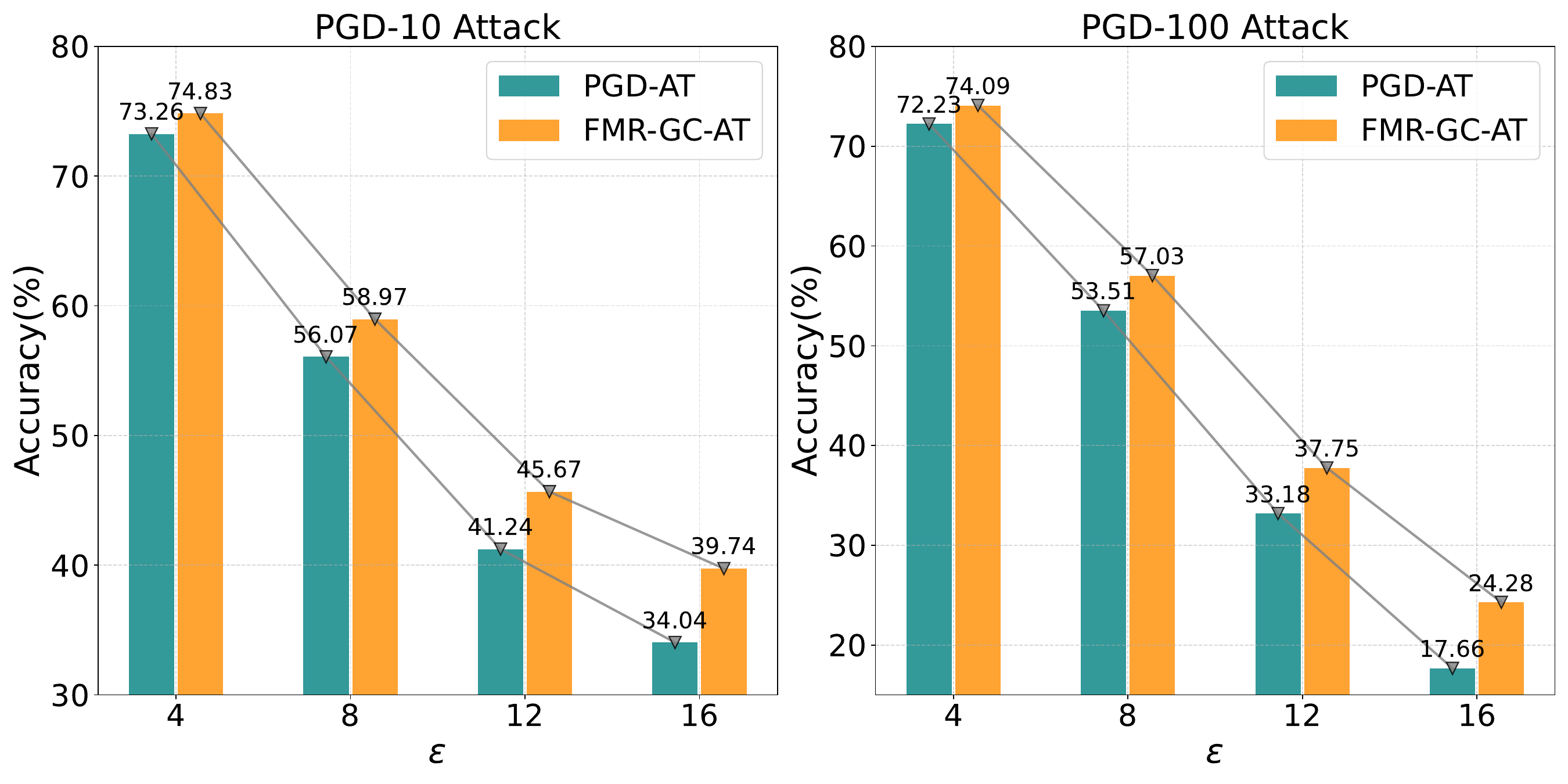}
    \end{center}
    \caption{Comparisons with varying $\epsilon$ values on CIFAR-10 using WRN34-10. The $x$-axis represents the $\epsilon$ value, while $y$-axis represents the robust accuracy~(\%).} 
    \label{epison_ana}
\end{figure}

\par 1) \textbf{FMR-GC enhances the model's robustness against diverse amplitude adversarial attacks.} Our proposed method demonstrates improved performance when subjected to C\&W and PGD-10 attacks at different values of $\epsilon$. For example, under the PGD-10 attack, FMR-GC-AT showed respective increases of 1.57\%, 2.90\%, 4.43\%, and 5.70\% compared to PGD-AT at $\epsilon$ values of 4, 8, 12, and 16.
\par 2) \textbf{FMR-GC improves the generalization against attacks.} As $\epsilon$ increases, the model equipped with FMR-GC exhibits enhanced generalization capabilities by experiencing a smaller decrease in accuracy compared to the baseline approach. For instance, when $\epsilon$ increases from 8 to 16, PGD-AT accuracy decreased by 15.52\%, whereas FMR-GC-AT only observed a 13.47\% decline under the PGD-100 attack.

\begin{table}[t]
    \caption{Ablation studies on the robustness (accuracy (\%)) with and without different HMR-GC components against various adversarial attacks on CIFAR10 using ResNet-18. The \textbf{number} in bold indicates the best accuracy.}
    \label{Different_components}
    \begin{center}
    \resizebox{\columnwidth}{!}
    {
    \begin{tabular}{c|ccccc}
        \toprule
        Method & Clean & FGSM & PGD-10 & C\&W & AA\\ 
        \midrule
        ResNet-18 &80.18&74.30&51.92&48.90&47.10\\
        +$\tilde{D}^{-\frac{1}{2}}\tilde{A}\tilde{D}^{-\frac{1}{2}}$&81.03&75.93&54.12&50.30&49.39\\
        +$\theta_2$&\textbf{81.35}&\textbf{76.52}&\textbf{54.91}&\textbf{50.86}&\textbf{49.84}\\
        \bottomrule
    \end{tabular}
    }
    \end{center}
\end{table}

\subsection{Ablation Study on Impact Assessment of FMR-GC Modules}
In this part, we performed an evaluation study to assess the individual contributions of different components of the FMR-GC modules in enhancing the model's robustness. \Cref{Different_components} depicts the results derived from the integration of the Laplacian Matrix $\tilde{D}^{-\frac{1}{2}}\tilde{A}\tilde{D}^{-\frac{1}{2}}$ and a learnable weight matrix $\theta_2$ into the FMR-GC module. Our analysis leads us to two main conclusions:
\par 1) \textbf{Laplacian matrix operations enhance the robustness of the obtained representations.} 
Through the simple multiplication of the Laplacian matrix, $\tilde{D}^{-\frac{1}{2}}\tilde{A}\tilde{D}^{-\frac{1}{2}}$, with feature $X$ during the graph convolution process in \cref{GCN_Eq} during inference, FMR-GC notably improves the model's performance. For instance, under PGD-10 and AutoAttack attacks, the performance improved by 2.20\% and 2.29\% respectively, while the accuracy on clean samples increased by 0.85\%. 
The Laplacian matrix serves to aggregate and propagate neighborhood features of nodes, thereby enabling the extraction of richer features. Furthermore, it takes into account the contextual information of spatial positions during the feature propagation process, employing neighborhood features to assist in the calibration of perturbed features.
These factors contribute to a substantial enhancement in model performance when facing attacks, while maintaining accuracy on clean samples.
\par 2) \textbf{Incorporating learnable weight matrices further improves model performance.}
The introduction of a learnable weight matrix $\theta_2$ in the training process bolsters the model's performance beyond just using the Laplacian matrix. For example, when subjected to FGSM and C\&W attacks, the model's performance exhibits enhancements of 0.59\% and 0.56\% respectively. This optimization results from the adaptive nature of the weight matrix, which empowers the FMR-GC to amend the weights dynamically, thereby intensifying the model's aptitude to comprehend the inter-node relations and significance while processing node data.

\begin{table}[t]
    \begin{center}
     \caption{
        Comparison of computational costs (\# Params and Times) between the original model and our approach on CIFAR-10.
    }
    \label{tb:computation}
    \resizebox{\columnwidth}{!}
    {
        \begin{tabular}{c|c c|c c}
            \hline
            \multicolumn{1}{c|}{\multirow{2}{*}{Method}} & \multicolumn{2}{c|}{\textit{\textbf{WRN34-10}}} & \multicolumn{2}{c}{\textit{\textbf{ResNet-18}}}\\
            \cline{2-5}
             ~ & \# Params (M) & Flops (G) & \# Params (M) & Flops (G)\\
            \hline
            Original     &46.16&6.670&11.17&1.113\\
            + FMR-GC     &48.25&6.737&13.26&1.247\\
            \hline
        \end{tabular}}
    \end{center}
\end{table}

\subsection{Analysis of Computational Cost}
\Cref{tb:computation} presents a comparison of our approach and the original model in terms of training efficiency. This comparison includes an analysis of the model parameters(\# params (M)) and the number of floating point operations (FLOPs (G)). The results highlight that our model enhances the robustness of the model without a significant increase in model parameters or the cost of AT.
\par
For example, after the integration of the FMR-GC module, both models exhibited an increase in parameter count and computational complexity. In the case of WRN34-10, floating point operations increased by approximately 0.067 G. Similarly, for ResNet-18, floating point operations recorded an increase of about 0.134 G. This indicates that the introduction of the FMR-GC module indeed enhances the complexity and computational overhead of the models. However, it is important to note that this additional complexity and computational cost have proven to be justified, as evidenced by the significant improvement in model performance.

\section{Conclusion}
In this study, we introduce a novel plug-and-play module called Feature Map-based Reconstructed Graph Convolution (FMR-GC). This module harmonizes the feature maps and contextual information during the network inference process to calibrate contaminated feature activations to a normal state. 
To tackle this objective, we initially regard each feature map extracted from a convolutional layer during network inference as a node and execute graph reconstruction in the channel dimension. Following this, we employ graph convolution operations to capture contextual information of the nodes, thereby utilizing neighborhood features to help calibrate contaminated feature activations.
The FMR-GC module is flexible and compatible with different adversarial training methods. Extensive experiments have been conducted to demonstrate the superiority of our proposed approach.


\section*{Acknowledgements}
This work is supported by the National Natural Science Foundation of China (No.~62276221, No.~62376232), and the Science and Technology Plan Project of Xiamen (No. 3502Z20221025), the Open Project Program of Fujian Key Laboratory of Big Data Application and Intellectualization for Tea Industry, Wuyi University (No.~FKLBDAITI202304).

\bibliographystyle{elsarticle-harv}\biboptions{authoryear}
\bibliography{reference}

\end{document}